\documentclass[letterpaper]{article} % DO NOT CHANGE THIS
\usepackage{aaai25}  % DO NOT CHANGE THIS
\usepackage{times}  % DO NOT CHANGE THIS
\usepackage{helvet}  % DO NOT CHANGE THIS
\usepackage{courier}  % DO NOT CHANGE THIS
\usepackage[hyphens]{url}  % DO NOT CHANGE THIS
\usepackage{graphicx} % DO NOT CHANGE THIS
\usepackage{natbib}  % DO NOT CHANGE THIS AND DO NOT ADD ANY OPTIONS TO IT
\usepackage{caption} % DO NOT CHANGE THIS AND DO NOT ADD ANY OPTIONS TO IT
\usepackage{algorithm}
\usepackage{algorithmic}
\usepackage{multirow}
\usepackage{utfsym}
\usepackage{color}
\usepackage{amssymb}
\usepackage{enumitem}
\usepackage{amssymb} % For checkboxes
\usepackage{newfloat}
\usepackage{listings}
\DeclareCaptionStyle{ruled}{labelfont=normalfont,labelsep=colon,strut=off} % DO NOT CHANGE THIS
\floatstyle{ruled}
\newfloat{listing}{tb}{lst}{}
\floatname{listing}{Listing}
\title{PDDM: Pseudo Depth Diffusion Model for RGB-PD Semantic Segmentation Based in Complex Indoor Scenes}
\author{
    Xinhua Xu, Hong Liu\thanks{Corresponding author: hongliu@pku.edu.cn}, Jianbing Wu, Jinfu Liu
}
\affiliations{
    State Key Laboratory of General Artificial Intelligence, Peking University, Shenzhen Graduate School\\
    \{xuxinhua, kimbing.ng, liujinfu\}@stu.pku.edu.cn, hongliu@pku.edu.cn
}

\usepackage{bibentry}
\begin{document}

\maketitle

\begin{abstract}
The integration of RGB and depth modalities significantly enhances the accuracy of segmenting complex indoor scenes, with depth data from RGB-D cameras playing a crucial role in this improvement. However, collecting an RGB-D dataset is more expensive than an RGB dataset due to the need for specialized depth sensors. Aligning depth and RGB images also poses challenges due to sensor positioning and issues like missing data and noise. In contrast, Pseudo Depth (PD) from high-precision depth estimation algorithms can eliminate the dependence on RGB-D sensors and alignment processes, as well as provide effective depth information and show significant potential in semantic segmentation. Therefore, to explore the practicality of utilizing pseudo depth instead of real depth for semantic segmentation, we design an RGB-PD segmentation pipeline to integrate RGB and pseudo depth and propose a Pseudo Depth Aggregation Module (PDAM) for fully exploiting the informative clues provided by the diverse pseudo depth maps. The PDAM aggregates multiple pseudo depth maps into a single modality, making it easily adaptable to other RGB-D segmentation methods. In addition, the pre-trained diffusion model serves as a strong feature extractor for RGB segmentation tasks, but multi-modal diffusion-based segmentation methods remain unexplored. Therefore, we present a \textbf{Pseudo Depth Diffusion Model (PDDM)} that adopts a large-scale text-image diffusion model as a feature extractor and a simple yet effective fusion strategy to integrate pseudo depth. To verify the applicability of pseudo depth and our PDDM, we perform extensive experiments on the NYUv2 and SUNRGB-D datasets. The experimental results demonstrate that pseudo depth can effectively enhance segmentation performance, and our PDDM achieves state-of-the-art performance, outperforming other methods by +6.98 mIoU on NYUv2 and +2.11 mIoU on SUNRGB-D.
\end{abstract}

 \begin{links}
	 \link{Code}{https://github.com/Oleki-xxh/PDDM}
\end{links}

\section{Introduction}

Semantic segmentation constitutes a significant task in computer vision, aiming to categorize each pixel into predefined categories. This technique is essential for various applications \cite{autonomous2,robots,robot}, such as autonomous driving, SLAM, and scene understanding, which has made significant strides (e.g., \cite{FCN,segformer}) in recent years. Nonetheless, in complex and dense indoor scenes, relying solely on the RGB modality is insufficient for accurate segmentation. To address this challenge, the depth modality, which provides valuable information on object positions, edges, and geometric information, is utilized to enhance segmentation accuracy, known as RGB-D semantic segmentation.

\begin{figure}
	\includegraphics[width=\linewidth]{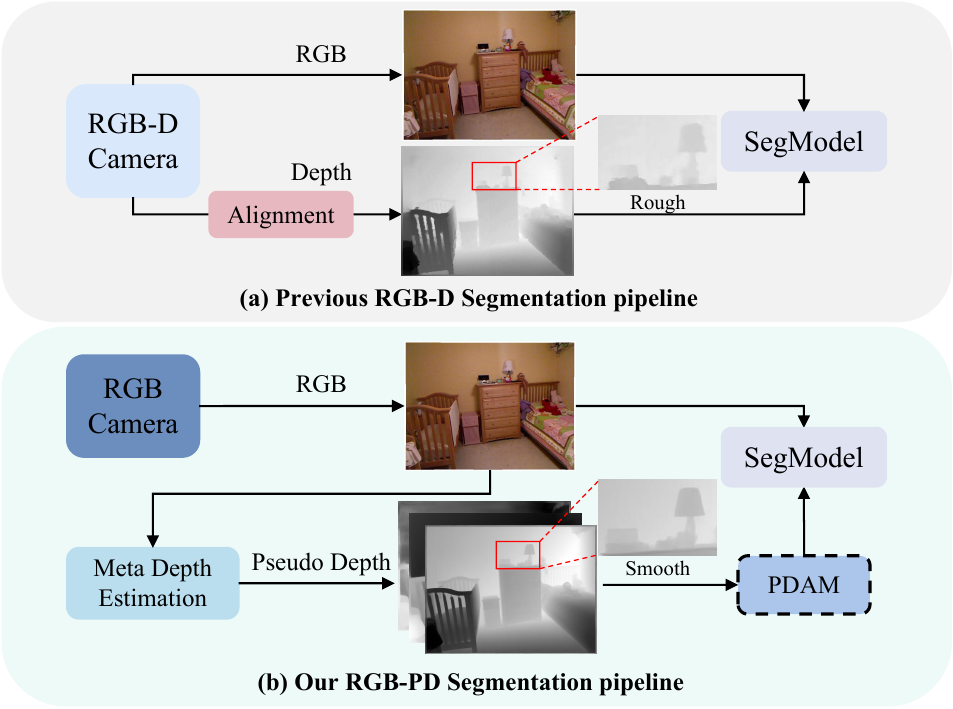}
	\caption{(a) Previous RGB-D segmentation methods utilize RGB and depth images from RGB-D cameras. (b) Our RGB-PD segmentation adopts pseudo depth images from RGB images by meta depth estimation methods and integrates them into a single modality by pseudo depth aggregation module (PDAM). It can be observed that the real depth image exhibits some noise, resulting in rougher edges, whereas the edges in pseudo depth images are comparatively smoother.}
	\label{fig1}
\end{figure}

Figure \ref{fig1} (a) shows the workflow of previous RGB-D methods. As can be observed, mainstream RGB-D semantic segmentation works \cite{CMX,delivering,tokenfusion} use both depth and RGB modalities from RGB-D cameras and emphasize the development of superior network architectures or fusion mechanisms. Although these methods achieve impressive results on various datasets, some notable limitations persist: Collecting an RGB-D dataset is more costly than an RGB dataset due to the need for specific depth sensors. Additionally, aligning raw depth and RGB images requires projecting them into the same coordinate space due to the different positions of the sensors \cite{nyuv2}. Furthermore, raw depth images often contain missing or erroneous holes and regional gaps due to sensor noise, object reflections, and shadow effects, as shown in Figure \ref{fig1} (a). Overall, these issues limit the utilization of RGB-D images.

Thereby, one question arises: Are RGB-D sensors really necessary for obtaining depth information? With the rapid advancement of monocular depth estimation \cite{depthanything,depth1}, the precision of depth prediction from a single RGB image has been significantly improved. We will show that Pseudo Depth (PD), predicted by depth estimation algorithms, possesses significant potential for improving semantic segmentation performance. Utilizing pseudo depth not only eliminates the reliance on depth sensors but also omits the alignment process. Additionally, as illustrated in Fig. \ref{fig1} (b), although the pseudo depth values may not be as accurate as real depth values, the region, edge, and geometric information can be well-captured by pseudo depth. It can be seen that the pseudo depth images are smoother and less noisy, without the instability often observed in real depth data because depth estimation methods are usually pre-trained on larger datasets, which is consistent with the current trend in the community. These points motivate us to build an \textbf{RGB-PD segmentation pipeline} to explore the practicality of utilizing pseudo depth instead of real depth for semantic segmentation, as illustrated in Figure \ref{fig1} (b). We utilize meta depth estimation algorithms to generate one or more pseudo depth images from a single RGB image.

Conventionally, pseudo depth is assumed to be less reliable than real depth. However, we prove that combining multiple pseudo depth predictions from various depth estimation methods can fully leverage their unique and valuable information, potentially matching or even exceeding the segmentation performance achieved with real depth. We propose the \textbf{Pseudo Depth Aggregation Module (PDAM)} that merges various pseudo depth maps into a unified pseudo depth map. This pseudo depth map, in conjunction with the RGB modality, is then fed into the SegModel, as shown in Figure \ref{fig1} (b). By exploiting spatial and channel correlations, PDAM allocates different weights to pseudo depth, thereby enhancing their feature representations and facilitating information complementation by exploiting the informative clues from various pseudo depth maps. Moreover, the proposed PDAM is plug-and-play and can be readily applied to existing RGB-D networks by replacing the real depth with aggregated pseudo depth, achieving comparable or even superior results, as demonstrated in the subsequent ablation studies.

On the other hand, the advent of diffusion models \cite{DDPM,stable} has given rise to a series of semantic segmentation methods \cite{odise,meta-prompt,label-efficient} that use large-scale diffusion models as backbones to leverage the rich high-level semantic information contained in their internal representations \cite{label-efficient}. However, integrating pseudo depth into the diffusion model for semantic segmentation remains unexplored. Since the data format of pseudo depth is the same as RGB, it is usually necessary to add a branch to extract pseudo depth features and achieve feature alignment. However, the extra parameters and computation costs caused by adding a whole diffusion branch are unbearable, which becomes a significant challenge.

Therefore, to address the above challenges and effectively harness pseudo depth data, we propose the \textbf{Pseudo Depth Diffusion Model (PDDM)}, an RGB-PD semantic segmentation model based on a large-scale text-image diffusion model to utilize information from RGB and pseudo depth modalities simultaneously. An overview of our approach is illustrated in Figure \ref{fig2}. As can be observed, PDDM obtains the aggregated pseudo depth through meta depth estimation methods and PDAM, and compresses the RGB image into latent representation to reduce computational complexity by utilizing the frozen pre-trained VAE \cite{autoenodelkl} encoder. With both inputs, a pre-trained text-to-image diffusion UNet model is employed to extract internal features. Additionally, features from the VAE are also utilized. With these features as input, the SegHead generates final segmentation results. To introduce pseudo depth into the PDDM, we propose to treat the pseudo depth features compressed by the pseudo depth encoder as structured noise, combine them with RGB features, and feed them into the diffusion model, aiming to fuse the pseudo depth information and enhance the model's perception of pseudo depth.

To validate the effectiveness of PDDM and pseudo depth, we conduct extensive experiments on two challenging datasets, NYUv2 and SUNRGB-D. Our approach achieves state-of-the-art performance on both datasets. Furthermore, we show the effectiveness of pseudo depth, and our PDAM is general and applicable to other RGB-D semantic segmentation methods, achieving performance improvement.

In summary, our contributions are as follows:

\begin{figure*}
	\centering
	\includegraphics[width=0.85\textwidth]{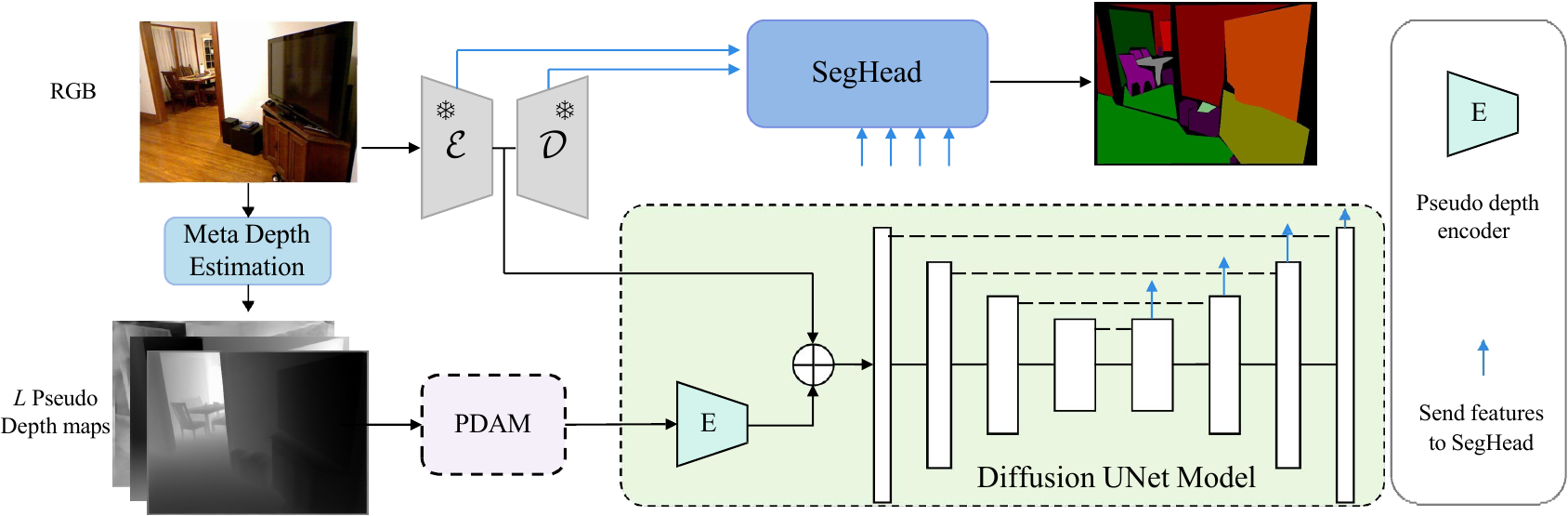}
	\caption{The overall architecture of the proposed PDDM. Given an RGB image, we first predict $L$ pseudo depth maps through Meta Depth Estimation methods and integrate them into a single aggregated pseudo depth with PDAM. Then, the RGB image is compressed to the latent features by the frozen VAE encoder $\mathcal{E}$. With both inputs, a pre-trained text-to-image diffusion UNet model is employed to extract internal features. With the features from diffusion UNet and VAE as inputs, the SegHead generates the final segmentation results. 
	}
	\label{fig2}
\end{figure*}

\noindent
$\bullet$ We pioneer an RGB-Pseudo Depth semantic segmentation pipeline with PDAM to explore the practicality of utilizing pseudo depth, where PDAM is plug-and-play and can be easily applied to other methods. 

\noindent
$\bullet$ We propose the Pseudo Depth Diffusion Model (PDDM),  an RGB-PD semantic segmentation model based on a large-scale text-image diffusion model to utilize information from RGB and pseudo depth modalities simultaneously, incorporating pseudo depth into the diffusion model by a simple yet effective fusion method.

\noindent
$\bullet$ In extensive experiments on the NYUv2 and SUNRGB-D datasets, RGB-PD segmentation pipeline and PDAM exhibit the effectiveness of pseudo depth. Our PDDM outperforms all existing baselines on both datasets, achieving state-of-the-art performance.

\section{Related Work}

\textbf{RGB-D Semantic Segmentation}. To facilitate interaction and alignment between RGB and depth modalities, most methods make a considerable effort to develop fusion modules. For example, CMX \cite{CMX} utilizes the depth modality through multi-level cross-modal interactions. DFormer \cite{dformer} introduces an RGB-D segmentation backbone designed to fully extract and utilize the 3D geometric information encapsulated in the depth modality. Undeniably, the designs of these interaction and fusion mechanisms have significantly enhanced performance. However, all of these approaches overlook a critical aspect: depth information can be derived not only from RGB-D sensors but also from high-precision depth estimation methods. Although some methods \cite{depthsemantic,depthsemantic1,depthsemantic2} incorporate depth estimation modules into semantic segmentation networks to leverage depth estimation features, thereby enhancing semantic segmentation performance. However, these approaches exhibit a strong dependency on the network's design and still require real depth data to supervise the training phase. Consequently, they are not readily applicable to other methods. Additionally, pseudo depth is also widely utilized in salient object detection. CDNet \cite{cdnet} combines real depth and pseudo depth to reduce the noise of depth information, but remains dependent on real depth. MGSNet \cite{MGSNet} and PopNet \cite{PopNet} derive pseudo depth through various methods but only utilize a single pseudo depth map, leading to insufficient depth robustness. In contrast, our proposed RGB-PD pipeline facilitates the direct application of pseudo depth predicted by depth estimation methods, along with a pseudo depth aggregation module, to other approaches, thereby steadily enhancing performance.

\textbf{Diffusion-based Semantic Segmentation}. With the proposal of Denoising Diffusion Probabilistic Models (DDPM) \cite{DDPM}, numerous studies have begun to explore the potential of diffusion models in semantic segmentation. Existing diffusion-based semantic segmentation methods can be divided into two categories: (1) The first category formulates the segmentation task as a general conditional denoising process, generating the segmentation result from a Gaussian noise map with an RGB image as the condition \cite{DDP,segdiff,medsegdiff}. While these works verify that the diffusion model can be applied to the segmentation task, they require a long time for inference since the diffusion process requires multiple diffusion steps. (2) The second type of approach uses the pre-trained diffusion model as a feature extractor and feeds the features into the segmentation head to generate results. For example, ODISE \cite{odise} proposes using a large-scale text-to-image diffusion model \cite{stable} for the open-vocabulary segmentation task, incorporating a learnable implicit text embedding. MetaPrompt-SD \cite{meta-prompt} presents a method to use a limited set of meta prompts to fine-tune the diffusion model for the visual perception task. Owing to the strong performance of the diffusion model, these studies effectively harness its exemplary capabilities, significantly enhancing semantic segmentation performance. However, these approaches have exclusively utilized the RGB modality, overlooking the integration of supplementary information from other modalities, such as depth. Building on the second type of method, we propose to adeptly combine both pseudo depth and RGB modalities to achieve more comprehensive and accurate semantic segmentation.

\section{Methodology}

\subsection{Pseudo Depth Diffusion Model}

Figure \ref{fig2} illustrates the overall architecture of our PDDM. Given an RGB image, it is first passed through the Meta Depth Estimation methods to obtain $L$ pseudo depth images, whereas Meta Depth Estimation can be multiple arbitrary depth estimation methods. After that, the generated pseudo depth images are fed into the Pseudo Depth Aggregation Module (PDAM) for integration into a single pseudo depth modality (applicable only for multiple pseudo depth images). This single pseudo depth modality contains effective information from the pseudo depth images.

On the other hand, the input RGB image is processed by the frozen encoder $\mathcal{E}$ and decoder $\mathcal{D}$ of the VAE \cite{autoenodelkl} for image compression, which reduces the resolution of the RGB image to 1/8 of its original size, yielding RGB latent features. Then, we employ a pre-trained text-to-image diffusion model with time step $t$ as a feature extractor for semantic segmentation, and the RGB latent features and aggregated pseudo depth are fed into the Diffusion UNet Model, performing a single forward for feature extraction. Following \cite{meta-prompt}, we do not add noise to the RGB latent features. In addition, we use learnable text embedding sent to diffusion UNet in cross-attention to guide feature extraction. We finally select the output features from VAE and the diffusion UNet decoder, and then send the integrated features to the SegHead to generate segmentation results. For the fusion of RGB and pseudo depth modality, we propose to fuse the pseudo depth information in a simple yet effective way.

\begin{figure}
	\centering
	\includegraphics[width=0.90\linewidth]{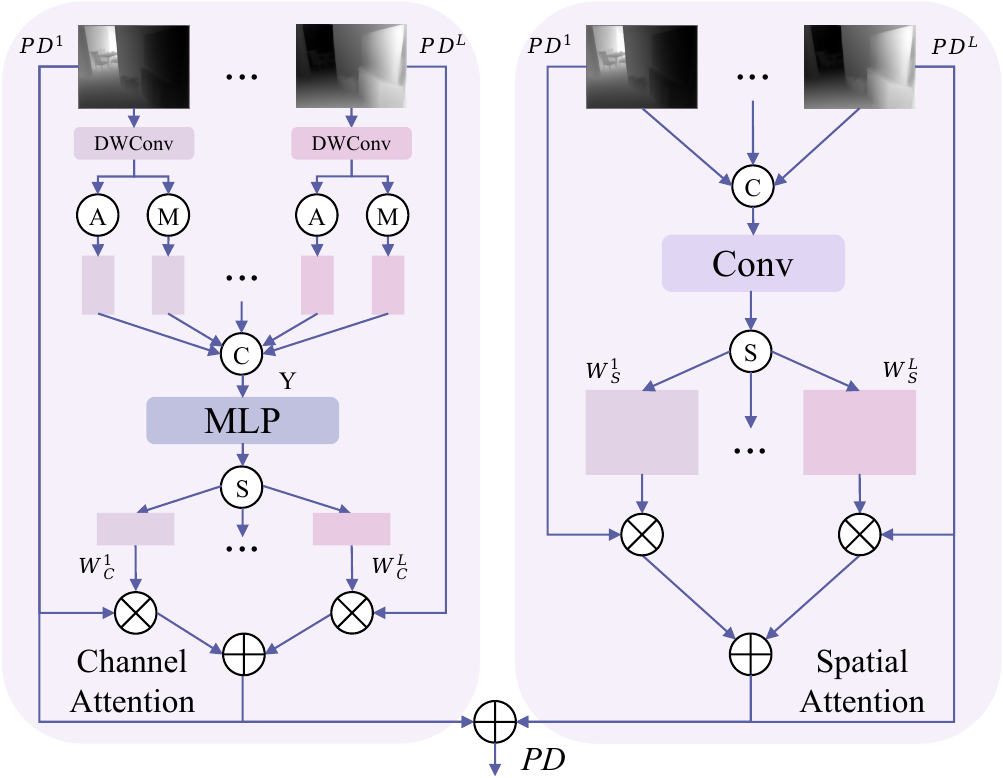}
	\caption{The structure of PDAM. It aims to aggregate effective information from multiple pseudo depth maps.}
	\label{fig3}

\end{figure}

\subsection{Pseudo Depth Aggregation Module}
Pseudo depth can be obtained from various depth estimation methods, and the effective information contained within the pseudo depth estimated by different methods varies. Although methods with higher accuracy tend to yield pseudo depth that contains more valid information, low-quality pseudo depth map also possesses unique insights. The useful information can be aggregated by using features coming from different pseudo depth maps. To this purpose, we propose the Pseudo Depth Aggregation Module (PDAM) as shown in Figure \ref{fig3}.

Suppose we obtain $L$ pseudo depth images, each with three channels denoted as $PD=\{PD^i \in \mathbb{R}^{H\times W\times 3}\}_{i=1}^{L}$, where $H$ and $W$ represent the height and width of an image, respectively. PDAM, which is divided into the channel attention stage and the spatial attention stage, exploits the correlation between channel and spatial dimensions, integrating the informative clues from various pseudo depth maps into a single pseudo depth, thus achieving the complementation and aggregation of information and enhancing the overall quality and usability of pseudo depth.

During the channel attention phase, we first conduct a 5$\times$5 depth-wise convolution and global max pooling and global average pooling on each pseudo depth input to compress information. These operations produce one-dimensional vectors with a length equal to the channel dimension, which is concatenated and fed through a Multi-Layer Perceptron (MLP), resulting in multiple sets of channel dimension weights ${W}_{{C}}^{{i}} \in \mathbb{R}^{1\times 1\times 3}$. These weights are then split and multiplied with the corresponding pseudo depth inputs in the channel dimension to modulate the channel-wise features as follows:

\begin{equation}
	{W}_{{C}}^{{1}},...,{W}_{{C}}^{{L}}=Split\bigl(MLP({Y})\bigr),
\end{equation}
\begin{equation}
	{F}_{{C}}^{{i}}={W}_{{C}}^{{i}} \otimes {PD}^{{i}}, 
\end{equation}
where $Y$ denotes the concatenated vector and ${F}_{{C}}^{{i}}$ denotes the channel attention features of ${PD}^{{i}}$. The $MLP(\cdot)$ contains two linear projection layers, a ReLU function and a sigmoid function.

\begin{table*}[htbp]
	\begin{center}
		\scalebox{0.88}{
			\begin{tabular}{lllllcccccc}
				\hline
				\multirow{2}{*}{Method}&\multirow{2}{*}{Publication}&\multirow{2}{*}{Modality}&\multirow{2}{*}{MS}&\multirow{2}{*}{Backbone}&\multicolumn{3}{c}{NYUv2}&\multicolumn{3}{c}{SUNRGB-D}\\
				\cline{6-11}
				~&~&~&~&~&PA&MA&mIoU&PA&MA&mIoU\\
				\hline\hline
				EMSANet \cite{EMSANet} &IJCNN2022&RGB&&ResNet-34&73.02&58.25&44.42&80.13&55.83&44.52\\
				CMX \cite{CMX} &T-ITS2023&RGB&\checkmark&MiT B2&75.79&61.94&48.97&80.67&56.45&45.89\\
				CMNext \cite{delivering} &CVPR2023&RGB&\checkmark&MiT B4&77.68&65.89&52.72&81.45&57.56&47.35\\
				DFormer \cite{dformer} &ICLR2024&RGB&\checkmark&DFormer-S&77.12&65.07&51.40&81.74&59.31&48.14\\
				ODISE \cite{odise}&CVPR2023&RGB& &SD&82.19&71.88&59.84&83.13&62.10&51.20\\
				\hline
				SA-Gate \cite{SA-Gate}&ECCV2020&RGB-D&\checkmark&ResNet-101&77.9&-&52.4&-&-&49.4\\
				SGNet \cite{SGNet}&TIP2021&RGB-D&\checkmark&ResNet-101&76.8&63.3&51.1&82.0&60.7&48.6\\
				CEN \cite{CEN}&TPAMI2022&RGB-D&\checkmark&ResNet-152&77.7&65.0&52.5&83.5&63.2&51.1\\
				ShapeConv \cite{ShapeConv}&ICCV2021&RGB-D&\checkmark&ResNext-50&76.4&63.5&51.3&82.2&59.2&48.6\\
				PGDENet \cite{pgdenet}&TMM2022&RGB-D& &ResNet-34&78.1&66.7&53.7&$\textbf{\textcolor{blue}{87.7}}$&61.7&51.0\\
				FRNet \cite{FRNet}&J-STSP2022&RGB-D& &ResNet-34&77.6&66.5&53.6&87.4&62.2&51.8\\
				TokenFusion \cite{tokenfusion}&CVPR2022&RGB-D& &MiT-B3&79.0&66.9&54.2&84.7&64.1&53.0\\
				EMSANet \cite{EMSANet}&IJCNN2022&RGB-D& &ResNet-34&76.72$^{\dag}$&64.79$^{\dag}$&51.0&81.83$^{\dag}$&60.96$^{\dag}$&48.4\\
				MultiMAE \cite{MultiMAE}&ECCV2022&RGB-D& &ViT-B&-&-&56.0&-&-&$51.1$\\
				Omnivore \cite{omnivore}&CVPR2022&RGB-D& &Swin-B&-&-&54.0&-&-&-\\
				CMX \cite{CMX}&T-ITS2023&RGB-D&\checkmark&MiT-B5&80.1&-&56.9&83.8&-&52.4\\
				CMNext \cite{delivering}&CVPR2023&RGB-D&\checkmark&MiT-B4&79.78$^{\dag}$&69.96$^{\dag}$&56.9&-&-&51.9\\
				DFormer \cite{dformer}&ICLR2024&RGB-D&\checkmark&DFormer-L&80.25$^{\dag}$&70.47$^{\dag}$&57.2&83.83$^{\dag}$&64.04$^{\dag}$&52.5\\
				\hline
				PDDM (Ours)&-&RGB-PD& &SD&83.83&75.36&63.60&84.52&67.47&54.44\\
				PDDM (Ours)&-&RGB-PD&\checkmark&SD&$\textbf{\textcolor{blue}{84.10}}$&$\textbf{\textcolor{blue}{75.82}}$&$\textbf{\textcolor{blue}{64.18}}$&84.82&$\textbf{\textcolor{blue}{67.79}}$&$\textbf{\textcolor{blue}{55.11}}$\\
				\hline
				
		\end{tabular}}
		
	\end{center}
	\caption{Comparison with state-of-the-art methods on NYUv2 and SUNRGB-D. MS: Multi-Scale test. $^{\dag}$ indicates we reproduce the same results using the official implementation and checkpoints. }
	\label{table1}
	
\end{table*}

For the spatial attention stage, pseudo depth images are first concatenated and then subjected to a convolution, resulting in multiple sets of spatial weights maps ${W}_{{S}}^{{i}} \in \mathbb{R}^{H\times W\times 1}$. These weights are then split, and each is multiplied with its corresponding pseudo depth input in the spatial dimension. The process is expressed as follows:

\begin{equation}
	{W}_{{S}}^{{1}},...,{W}_{{S}}^{{L}}=Split\bigl(Conv({PD}^1 \mathbin\Vert ... \mathbin\Vert {PD}^L)\bigr),
\end{equation}
\begin{equation}
	{F}_{{S}}^{{i}}={W}_{{S}}^{{i}} \otimes {PD}^{{i}}, 
\end{equation}
where $\mathbin\Vert$  denotes the concatenation and ${F}_{{S}}^{{i}}$ denotes the spatial attention features of ${PD}^{{i}}$. $Conv(\cdot)$ consists of two convolution layers with a kernel size of 1$\times$1. 

The final aggregated pseudo depth is obtained via:
\begin{equation}
	PD=\sum_{i=1}^{L}({PD}^{i}+ \lambda_C {F}_{{C}}^{{i}}+\lambda_S {F}_{{S}}^{{i}}), 
\end{equation}
where $\lambda_C$ and $\lambda_S$ are two hyperparameters, with both set to 0.5 by default. By integrating information from various pseudo depth inputs and assigning different weights to pixels across channels and spatial positions, PDAM can effectively exploit the informative clues from each pseudo depth input. Moreover, since PDAM only incorporates simple MLP and convolution operations, it is efficient and deployment-friendly, requiring minimal parameters and computational resources. With multiple pseudo depth maps, we can apply PDAM to other RGB-D segmentation methods by replacing the real depth input with aggregated pseudo depth.

\subsection{Fusion Strategy}

With regard to utilizing depth information for semantic segmentation, as mentioned above, there are many excellent works \cite{delivering,tokenfusion,dformer} that have proposed effective methods, but most of them are based on a dual-branch structure to process RGB and depth modality respectively. In our PDDM, the increased amount of parameters and computation costs caused by adding a whole branch is unbearable. Therefore, we propose a simple yet effective fusion strategy to introduce pseudo depth information into the diffusion model with only a slight increase in computation and parameters.

Specifically, in the original stable diffusion framework, given an RGB image, noise is incorporated into the RGB image via:

\begin{equation} 
	{Z} = \sqrt{\overline{\alpha}_t} Z_{RGB} + \sqrt{1 - \overline{\alpha}_t}\epsilon,  \epsilon \sim \mathcal{N}(0,\mathbf{I})
	%	{Z} = \alpha Z_{RGB} + (1 - \alpha)E(PD)
\end{equation}
where $Z_{RGB}$ denotes the RGB latent features derived from the VAE encoder, and $t$ is the diffusion step we use. $\alpha_1$,...,$\alpha_T$ represent a pre-defined noise schedule, where $\overline{\alpha}_T = \prod_{k=1}^{T}\alpha_k$ is determined by $t$, as defined in \cite{DDPM}. $\epsilon$ is a Gaussian noise map. To introduce pseudo depth information into the diffusion model, we employ a pseudo depth encoder to compress the pseudo depth to the same shape as $Z_{RGB}$ and perform the fusion by addition as follows:

\begin{equation} \label{eq6}
	{Z} = \sqrt{\overline{\alpha}_t} Z_{RGB} + \sqrt{1 - \overline{\alpha}_t}\Bigl(E(PD)\Bigr),
\end{equation}
where $E(\cdot)$ denotes the pseudo depth encoder, which consists of three convolutions with kernel size 3$\times$3, stride 2, and padding 1, and $t$ is the diffusion step we use. In the above formula, we treat the compressed pseudo depth as structured noise to replace the Gaussian noise $\epsilon$,  to make the diffusion model pay as much attention to pseudo depth as to noise for feature extraction. This approach has proven effective, as demonstrated in the ablation study.

\begin{table*}[htbp]
	
	\begin{center}

		\scalebox{0.88}{
			\begin{tabular}{ccccccccccc}
				\hline
				\multicolumn{4}{c}{Depth Source}&\multirow{2}{*}{Fusion}&{EMSANet}&{CMX}&{CMNeXt}&{DFormer}&\multirow{2}{*}{PDDM (Ours)}\\
				\cline{1-4}
				RD&DA&MPSD&Mari&~&(ResNet34)&(MiT B2)&(MiT B4)&(Dformer-S)&~\\
				\hline\hline
				~&~&~&~&~&44.42&48.97&52.72&51.40&62.05\\
				\checkmark&~&~&~&~&48.60$^{\dag}$&52.63$^{\dag}$&\textbf{54.39}$^{\dag}$&52.91$^{\dag}$&\textbf{63.10}\\
				~&\checkmark&~&~&~&47.71&52.43&53.54&52.24&62.80\\
				~&~&\checkmark&~&~&\textbf{49.21}&\textbf{53.54}&53.25&\textbf{53.72}&63.03\\
				~&~&~&\checkmark&~&47.56&52.42&53.61&53.36&62.67\\
				\hline
				~&\checkmark&\checkmark&\checkmark&addition&48.89&51.19&52.94&53.59&62.78\\
				~&\checkmark&\checkmark&~&PDAM&49.17&53.78&54.19&53.10&63.16\\
				~&\checkmark&~&\checkmark&PDAM&48.75&54.08&53.98&52.80&63.08\\
				~&~&\checkmark&\checkmark&PDAM&49.19&53.88&54.31&53.87&63.21\\
				~&\checkmark&\checkmark&\checkmark&PDAM&\textbf{49.38}&\textbf{54.16}&\textbf{54.89}&\textbf{54.49}&\textbf{63.60}\\
				\hline
				
		\end{tabular}}
		
	\end{center}
	\caption{Ablation study of pseudo depth and PDAM on NYUv2 dataset with diffƒerent methods. RD: Real depth; DA: Depth Anything; MPSD: Meta Prompt-SD; Mari: Marigold. $^{\dag}$ indicates our implemented results.}
	\label{table2}
	
\end{table*}

\section{Experiments}
\subsection{Implementation Details}
\textbf{Datasets \& Pseudo Depth Estimation. }We validate our approach through experiments on the NYUv2 \cite{nyuv2} and SUNRGB-D \cite{sunrgbd} datasets. NYUv2 contains 1,449 images of size 640 $\times$ 480, annotated in 40 classes. We use the standard split of 795 training images and 654 testing images. SUNRGB-D integrates multiple indoor datasets, including NYUv2, with 5,285 training images and 5,050 testing images across 37 classes. Following \cite{CMX}, we randomly crop and resize the input images to 480 $\times$ 480. For generating pseudo depth, we set $L=3$ and select three leading distinct algorithms: Depth Anything \cite{depthanything}, Meta Prompt-SD \cite{meta-prompt}, and Marigold \cite{marigold}.

\textbf{Training Details}. For the diffusion model, we adopt the pre-trained stable diffusion \cite{stable} as our text-to-image diffusion UNet model. The VAE also comes from the stable diffusion model. We set the diffusion time step $t$ to 0 and implement Mask2Former \cite{mask2former} as the segmentation head. We extract features from the middle blocks indexed \{5, 7\} of the VAE encoder and decoder, and \{2, 5, 8, 11\} of the diffusion UNet decoder, then resize them to create a feature pyramid, and send it to the segmentation head. For data augmentation, we employ random scaling, cropping, flipping, and slight color jitter. For supervision, we adopt the dice loss, mask loss, and cross-entropy loss. We select the AdamW optimizer for parameter optimization with a weight decay of 0.05, setting the initial learning rate for the diffusion UNet and the additional branch at 5e-6 and 1e-4 for the remaining parameters. The VAE is frozen. We train PDDM with a batch size of 2 for 60k and 100k iterations on NYUv2 and SUNRGB-D, respectively. The learning rate is decayed by a factor of 0.1 at 40k and 60k steps on NYUv2 and SUNRGB-D, respectively. We evaluate the proposed PDDM and existing SOTA methods in terms of pixel accuracy (PA), mean pixel accuracy (MA), and mean intersection over union (mIoU).

\subsection{Comparison with State-Of-The-Art Methods} 
Table \ref{table1} summarizes the extensive comparisons between our PDDM and other 13 recent RGB-D segmentation methods on NYUv2 and SUNRGB-D datasets. Meanwhile, we adapt ODISE \cite{odise} to these two datasets to evaluate its performance. As can be observed, our PDDM achieves 63.60 mIoU on NYUv2, better than the 59.84 mIoU achieved by ODISE, which also uses SD as the backbone, demonstrating the effectiveness of PDDM and pseudo depth. Even with single-scale inference, our PDDM already exceeds previous methods, achieving promising results with 63.60 mIoU on NYUv2 and 54.44 mIoU on SUNRGB-D. When applying the multi-scale inference strategy, our PDDM further improves the results to 64.18 mIoU and 55.11 mIoU, which significantly outperforms other models by a large margin: +6.98 mIoU on NYUv2, +2.11 mIoU on SUNRGB-D. \textbf{{Overall, our PDDM achieves state-of-the-art performance with a large margin across these two benchmark datasets using multiple pseudo depth maps.}}

\subsection{Ablation Study} 

\textbf{Pseudo Depth and PDAM.} 
To validate the effectiveness of pseudo depth and PDAM, we evaluate several recent RGB-D segmentation methods on NYUv2, including EMSANet \cite{EMSANet}, CMX \cite{CMX}, CMNeXt \cite{delivering}, DFormer \cite{dformer}, and our PDDM. As shown in Table \ref{table2}, we apply RGB-only, real depth (RD) and various pseudo depth maps generated by Depth Anything (DA) \cite{depthanything}, Meta Prompt-SD (MPSD) \cite{meta-prompt}, and Marigold (Mari) \cite{marigold} to these methods and report the mIoU results. As can be observed, the results demonstrate that pseudo depth significantly outperforms RGB-only inputs and performs comparably to real depth, highlighting that pseudo depth possesses valuable edge and geometric information for enhancing segmentation performance. We also observe that using pseudo depth from MPSD yields better results than using real depth in most situations, suggesting that although the depth value within pseudo depth exhibits discrepancies compared with the real depth, the quality of geometric, edge, and region information contained in pseudo depth sometimes can be higher than real depth.

\begin{table}[t]
	\begin{center}
		\scalebox{0.88}{
			\begin{tabular}{cccc}
				\hline
				RGB:PD&PA&MA&mIoU\\
				\hline\hline
				0.6:0.4&82.78&73.60&61.89\\
				0.8:0.2&83.33&74.67&62.91\\
				0.9:0.1&83.40&74.60&63.07\\
				0.95:0.05&83.50&74.84&63.34\\
				0.99:0.01&83.36&75.09&63.29\\
				1:0&83.23&73.97&62.05\\
				\hline
				1:0.03&\multirow{2}{*}{\textbf{83.83}}&\multirow{2}{*}{\textbf{75.36}}&\multirow{2}{*}{\textbf{63.60}}\\ 
				(structured noise)&~&~&~\\
				\hline
				
		\end{tabular}}
		
	\end{center}
	\caption{Ablation for modality weights.  }
	\label{table3}
\end{table}

Moreover, we explore the impact of integrating multiple pseudo depth images using different quantities of pseudo depth and fusion techniques. As illustrated in Table \ref{table2}, "addition" refers to fusing pseudo depth through simple arithmetic addition. It is observed that the aggregation of multiple pseudo depth maps generally results in better performance than using real depth, except when using simple addition to fuse pseudo depth. We infer that simple addition breaks the region and edge information, leading to the performance drop. Meanwhile, integrating 2 or 3 pseudo depth maps with PDAM also outperforms using only one pseudo depth, which indicates that pseudo depth maps generated by different methods contain unique and valuable information, and our PDAM is capable of aggregating them to achieve a more substantial performance enhancement compared to simple addition. Finally, using PDAM to aggregate 3 pseudo depth maps achieves the best results.

\textbf{Modality Weights.} In PDDM, we introduce pseudo depth as Eq. (\ref{eq6}), and the weights $\sqrt{\overline{\alpha}_t}$ is crucial for the performance. We first set the modality weights with different values manually and study the trends of different weights for integrating pseudo depth on NYUv2. From Table \ref{table3}, all metrics increase as weights decrease, with the best ratio being RGB:PD= 0.95:0.05. It means the smaller the weight of pseudo depth, the better, but not zero, indicating that an excessively large weight can break the RGB information. Regarding this phenomenon, we notice that when using $t$=0, the weight of noise (RGB:noise $\approx$ 1 : 0.03) added in the original process is close to this ratio. Therefore, we train our model using the compressed pseudo depth as structured noise, and the result is better, which demonstrates that the diffusion model pays great attention to the noise part. When we replace the noise with pseudo depth, the model also takes pseudo depth information into consideration as much as noise to extract features, thereby facilitating the information fusion and performance improvement.

\begin{figure}
	\centering
	\includegraphics[width=0.95\linewidth]{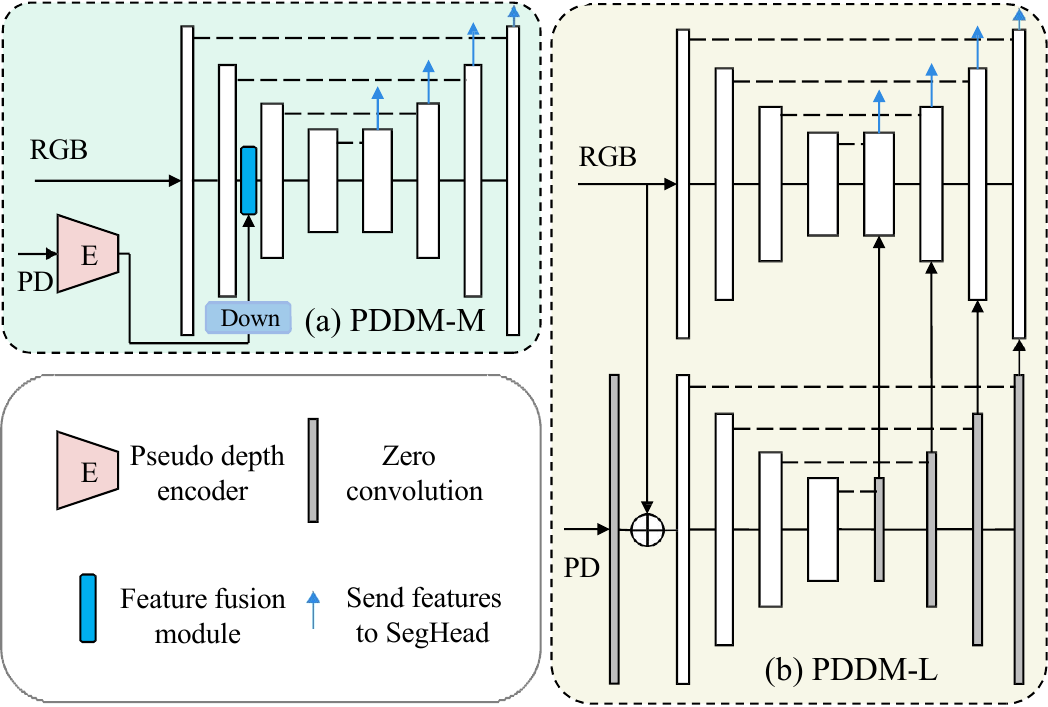}
	\caption{Framework of mid-term fusion (PDDM-M) and late fusion (PDDM-L).}
	\label{fig5}

\end{figure}

\begin{table}[t]
	
	\begin{center}
		\scalebox{0.88}{
			\begin{tabular}{cccc}
				\hline
				$t$&PA&MA&mIoU\\
				\hline\hline
				0&\textbf{83.83}&\textbf{75.36}&\textbf{63.60}\\
				50&83.32&74.74&62.77\\
				100&83.07&74.94&62.67\\
				200&83.00&74.27&62.50\\
				0+50+100&83.40&74.61&63.08\\
				\hline
				
		\end{tabular}}
		
	\end{center}

	\caption{Ablation for diffusion timestep $t$. }
	\label{table4}

\end{table}

\textbf{Diffusion Timesteps.} Based on using pseudo depth as structured noise, we also study which diffusion step(s) are most helpful for introducing pseudo depth into the diffusion model, where the results are shown in Table \ref{table4}. The performance decreases as t increases, and the best results are for $t$=0. Concatenating three steps, 0, 50, and 100, yields a performance drop compared to $t$=0 only, and requires 3 $\times$ inference time, further validating our optimal choice of $t$=0.

\textbf{Different Fusion Strategies.} 
Broadly speaking, our fusion strategy belongs to early fusion. In addition to early fusion, mid-term and late fusion are also commonly employed strategies. To further demonstrate the effectiveness of our fusion strategy, we conduct experiments to compare the performance of early, mid-term and late fusion. For clarity, we denote these three methods as PDDM-E, PDDM-M, and PDDM-L. For PDDM-M, as shown in Figure \ref{fig5} (a), we insert the feature fusion module into an intermediate layer of the diffusion UNet encoder to achieve feature-level fusion. We utilize the same pseudo depth encoder and downsampling to compress the pseudo depth and send the results to the feature fusion module in diffusion UNet encoder with RGB latent features. For feature fusion, we use CM-FRM \cite{CMX} by default. For PDDM-L, as shown in Figure \ref{fig5} (b), the RGB latent features are fed into the original diffusion UNet, and the pseudo depth is combined with RGB latent features to ensure the feature alignment and is subsequently fed into the additional branch to extract the pseudo depth features. In the decoder stage, each decoder feature is fed into the main branch by simple addition. Notably, the initial parameters of the encoder of the additional branch are copied from the main branch, and the decoder comprises only zero convolution to reduce parameters and computational costs.

\begin{table}[t]

	\begin{center}
		\scalebox{0.88}{
			\begin{tabular}{cccccc}
				\hline
				Strategy&mIoU&Extra parameters&Extra FLOPs\\
				%				RD&DA&MPSD&Mari&EMSANet(ResNet34)&CMX(MiT B2)&CMNeXt(MiT B4)&DFormer(S)&Ours\\
				\hline\hline
				Baseline&62.05&0M&0G\\
				PDDM-E&\textbf{63.60}&\textbf{0.37M}&\textbf{0.02G}\\
				PDDM-M&62.98&0.52M&0.63G\\
				PDDM-L&63.31&234.25M&361.23G\\
				PDDM-EML&61.80&234.77M&361.86G\\
				\hline
				
		\end{tabular}}
		
	\end{center}
	\caption{Comparison of PDDM with different strategy.  }
	\label{table7}

\end{table}

Table \ref{table7} presents the results of Baseline (without depth), PDDM-E, PDDM-M, PDDM-L, and PDDM-EML (apply early, mid-term and late fusion simultaneously) with default settings on NYUv2. We also detail the extra parameters and FLOPs required for incorporating pseudo depth into the diffusion model. In terms of performance comparison, PDDM-E, with 63.60 mIoU, showcases the highest efficiency, proving that early fusion is not only effective but also surpasses the more complex dual branch fusion strategy employed by PDDM-L. While PDDM-M struggles with aligning pseudo depth features with RGB features, PDDM-L achieves this alignment through an extra branch, enhancing its performance over PDDM-M and highlighting the critical role of feature alignment. In terms of computational efficiency, PDDM-E stands out, requiring only 0.37M extra parameters and 0.02G extra FLOPs with only a pseudo depth encoder, whereas PDDM-L requires significantly more extra parameters and FLOPs due to its additional branch. In addition, for PDDM-EML, it can be seen that introducing too much pseudo depth information in a non-unified manner not only fails to help the fusion of pseudo depth and RGB data but also incurs adverse effects, leading to a performance drop. Overall, the experimental results demonstrate that PDDM-E, which uses a simple yet effective manner to incorporate pseudo depth into the diffusion model, achieves optimal performance with minimal extra parameters and computation.

\section{Conclusions}
In this paper, we pioneer an RGB-PD semantic segmentation pipeline, featuring a pseudo depth aggregation module for extracting and refining valid information from diverse pseudo depth maps, which is plug-and-play and readily applicable to other methods. After aggregating multiple pseudo depth images, the performance is even better than that using real depth. In addition, we propose the Pseudo Depth Diffusion Model (PDDM) and investigate a simple yet effective fusion strategy to incorporate pseudo depth information into the diffusion model. Our PDDM outperforms all existing baselines on NYUv2 and SUNRGB-D datasets, establishing a new state-of-the-art performance.

\section{Acknowledgments}
This work was supported by National Natural Science Foundation of China (No.62373009).

\bibliography{aaai25}

\end{document}